\documentclass[letterpaper, 10 pt, conference]{ieeeconf}  % Comment this line out if you need a4paper

\IEEEoverridecommandlockouts                              % This command is only needed if 
\usepackage{amsmath} % assumes amsmath package installed
\usepackage{url}
\usepackage[dvipsnames]{xcolor}
\usepackage{pdfpages}
\usepackage{cite}
\usepackage{array}
\usepackage{upgreek}

\usepackage{dsfont}
\usepackage{amssymb}
\usepackage{wasysym}

\usepackage{booktabs}
\usepackage{dblfloatfix}    % To enable figures at the bottom of page

\definecolor{green(munsell)}{rgb}{0.0, 0.66, 0.47}

\title{\LARGE \bf
Light in the Larynx: a Miniaturized Robotic Optical Fiber for In-office Laser Surgery of the Vocal Folds
}

\author{Alex J. Chiluisa, Nicholas E. Pacheco,
Hoang S. Do, Ryan M. Tougas, Emily V. Minch,
Rositsa Mihaleva, \\ Yao Shen, 
Yuxiang Liu, Thomas L. Carroll, and Loris Fichera% <-this % stops a space
\thanks{This work was supported in part by the National Institutes of Health (NIH) under Award R15 DC018667 and in part by the National Science Foundation (NSF) under Award DGE-1922761. Any opinions, findings, and conclusions or recommendations expressed in this material are those of the authors and do not necessarily reflect the views of the NSF or NIH.}
\thanks{A.J. Chiluisa, R.M. Tougas, N.E. Pacheco,
and L. Fichera are with
the Department of Robotics Engineering, Worcester
Polytechnic Institute, Worcester,
MA 01609, USA (e-mail: ajchiluisa@wpi.edu)}% <-this % stops a space
\thanks{Y. Shen, H.S. Do, and Y. Liu are with the
Department of Mechanical Engineering, Worcester
Polytechnic Institute, Worcester,
MA 01609, USA}% <-this % stops a space
\thanks{E.V. Minch and R. Mihaleva are with the Department
of Biomedical Engineering, Worcester Polytechnic
Institute, Worcester, MA 01609, USA}% <-this % stops a space
\thanks{T.L. Carroll is with the Department of
Otolaryngology–Head and Neck Surgery, Harvard Medical School,
Boston, MA 02115, USA}}

\begin{document}

\maketitle
\thispagestyle{empty}
\pagestyle{empty}

%%%%%%%%%%%%%%%%%%%%%%%%%%%%%%%%%%%%%%%%%%%%%%%%%%%%%%%%%%%%%%%%%%%%%%%%%%%%%%%%
\begin{abstract}
This paper reports the design,
construction, and experimental validation
of a novel hand-held robot for in-office
laser surgery of the vocal folds.
In-office endoscopic laser surgery is an emerging
trend in Laryngology: It promises to deliver
the same patient outcomes of traditional
surgical treatment (i.e., in the operating room), at a fraction of the cost.
Unfortunately, office procedures can be 
challenging to perform; the optical fibers used 
for laser delivery can only emit light forward
in a line-of-sight fashion, which severely limits
anatomical access.
The robot we present in this paper aims to
overcome these challenges. 
The end effector of the robot is a 
steerable laser fiber, 
created through the combination of a thin
optical fiber (\diameter~0.225 mm) with a
tendon-actuated Nickel-Titanium 
notched sheath
that provides bending.
This device can be seamlessly used 
with most commercially available endoscopes,
as it is sufficiently small (\diameter~1.1 mm) to
pass through a working channel.
To control the fiber, we propose a
compact actuation unit that can be mounted
on top of the endoscope handle,
so that, during a procedure,
the operating physician can operate
both the endoscope and the steerable fiber
with a single hand.
We report simulation and phantom experiments
demonstrating that the proposed device
substantially enhances surgical access compared
to current clinical fibers.
\end{abstract}
\section{Introduction}
\label{sec:Introduction}
\begin{figure}\centering
    \includegraphics[width=1\linewidth]{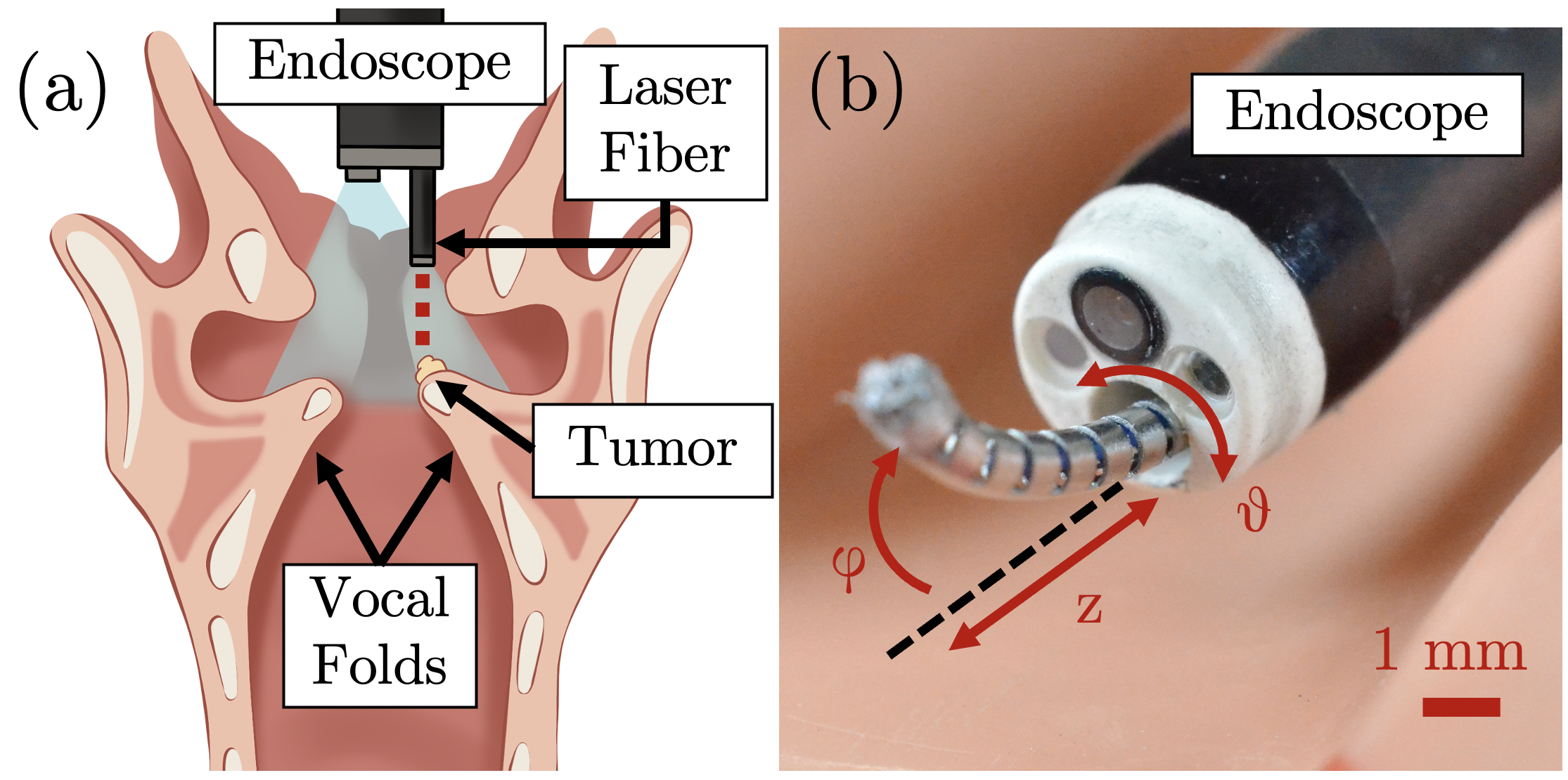}
  \caption{Office-based endoscopic laser surgery of the vocal folds.
  (a) Frontal cut of the human larynx illustrating the
  \textcolor{black}{current} surgical approach: an endoscope is deployed in proximity of the vocal folds and an optical fiber delivers laser pulses
  to the diseased tissue.
  The blue cone represents the field of view of the endoscopic camera, while the dashed red line represents the laser beam.
  The laser fiber cannot be controllably
  bent, which makes it impossible to
  treat tumors located off the fiber axis.
  Hard-to-reach regions include the two cavities
  immediately above the vocal folds and the inferior
  surface of the vocal folds.
  (b) Proposed device: a tendon-actuated
  miniaturized notched sheath made of Nickel-Titanium
  makes it possible to bend the optical fiber.
  In addition to bending, the sheath can also be
  rotated and translated, thus providing a 
  total of three degrees of freedom (DoFs).
}
  \label{fig:fig-1}
\end{figure}

%%5
In-office surgery is an
increasingly attractive option for the treatment of many
benign and pre-malignant tumors 
of the voice
box~\cite{Hantzakos2021}.
\textcolor{black}{In-office laser surgeries} 
are performed as illustrated in Fig.~\ref{fig:fig-1}:
a flexible channeled endoscope is passed into the larynx
by way of the \textcolor{black}{nasal cavity},
and
laser pulses
are applied on the diseased tissue until thermal
necrosis is achieved.
This approach represents a paradigm shift from how
laryngeal tumors are normally treated, and it
offers several important benefits.
First, office procedures typically only last a
few minutes and do not require general 
anesthesia.
Furthermore, because treatment is delivered
in the doctor's office, as opposed to the 
operating room, patient charges tend to be
substantially lower~\cite{Chen2021}.
Despite its documented benefits, in-office surgery is still
underutilized because of how challenging it can be to
perform.
One of the key limitations is the lack of
articulation in the optical fibers used
for laser delivery: it is only possible
to control the laser aiming indirectly, i.e.,
bending the endoscope's distal tip.
This greatly limits anatomical access~\cite{Chan2021},
and it also makes the procedure
disorienting for the operating
physician due to the inability to control
the laser aiming without also
moving the field of
vision\cite{DelSignore2016,Hu2017}.
Patients who present with a disease 
in hard-to-reach locations are not good candidates
for office treatment; and even if a procedure is
attempted, it can result in 
incomplete treatment, and therefore 
the need for additional follow-up
care~\cite{DelSignore2016}.
Seeking to overcome these limitations, in this paper,
we describe the design, construction, and validation of a new
robotic device to enable optical fiber bending during
endoscopic office procedures.
The device is shown in Fig. 1(b); it is
built by installing an off-the-shelf optical fiber into
a tendon-actuated Nickel-Titanium
continuum notched sheath.
The sheath diameter is 1.1 mm, which makes it
suitable for trans-luminal deployment through the 
operating channel of most
clinical endoscopes.
In addition to bending, the sheath can also be 
rotated and translated, thus providing a total of three degrees of freedom
(DoFs).
Actuation is provided by a modular add-on motor unit
that mounts on the endoscope handle.
We report experimental evidence, obtained in simulation and phantom experiments, documenting the ability of our
device to reach and deliver laser pulses to regions within
the larynx that are currently inaccessible in-office
procedures.
This paper is the first report on a surgical robotic device
specially designed
\textcolor{black}{to be compatible to use with a commercial endoscope}
for in-office laryngeal surgery.
\section{Background \& Related Work}
\label{sec:Related-Work}
Several other groups are actively developing
robots for laryngeal laser
surgery~\cite{Bajo2013,Acemoglu2019,Zhao2020,Mattos2021,Renevier2017,Kundrat2020,Fang2021},
but none of the existing prototypes can be 
readily used in the office procedures
we consider in this study.
In the following, we briefly describe the 
surgical setup and workflow of an office procedure.
We then illustrate the challenges that motivate
our work and formulate the specifications that
guided our design process.
\subsection{Overview of Office Procedures in the Larynx}
\begin{figure}\centering
    \includegraphics[width=1\linewidth]{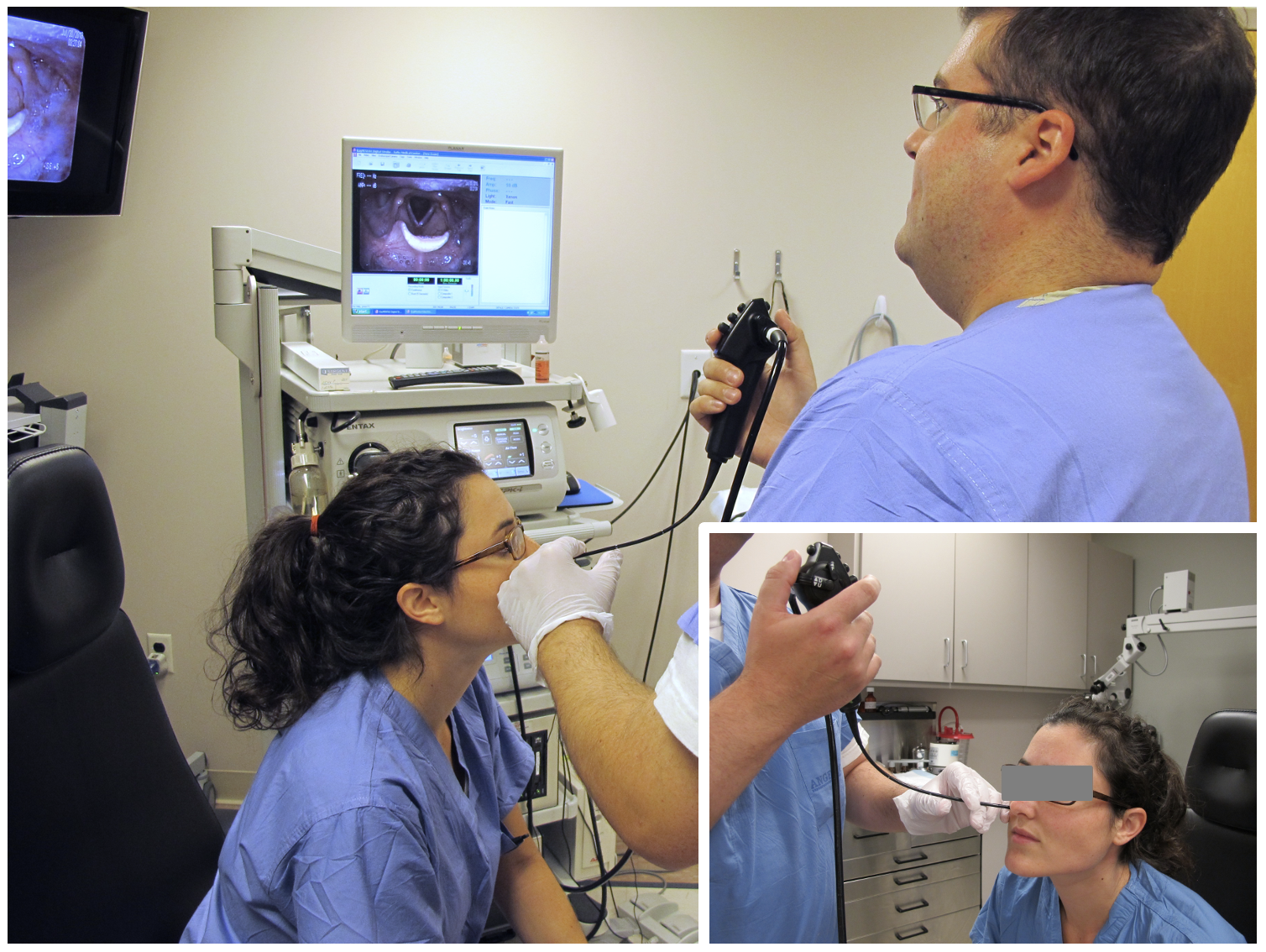}
  \caption{Setup for in-office laser surgery of the vocal
  folds. Patients receive the procedure awake sitting on
  an examination chair. The operating physician introduces
  a flexible endoscope into the larynx by way of the nose.
  Right-handed physicians typically hold the endoscope with
  their right hand; the left hand rests on the patient's 
  face to stabilize the endoscope and to
  feel any patient movements.}
  \label{fig:fig-2}
\end{figure}

Office-based laryngeal
procedures are performed as shown in 
Fig.~\ref{fig:fig-2}.
Patients receive the procedure awake while
sitting on an examination
chair.
To visualize the larynx, the operating physician
introduces a flexible endoscope through the nose,
prior to the administration of a topical 
anesthetic.
In principle, it would be possible
to pass the endoscope through the mouth, 
but this approach is generally avoided 
as it can easily trigger
the gag reflex~\cite{Zheng2021}.
The endoscope needs to be sufficiently 
small to pass through one of the nostrils, and 
ideally as small as possible to minimize
patient discomfort.
The typical endoscope diameter for these
procedures is 5 mm~\cite{Tibbetts2019,Hu2017}.
Endoscopes are equipped with a working channel,
typically 2 mm in diameter, which enables 
the passage of optical fibers for laser delivery.
The goal of a procedure is to thermally destroy 
diseased tissue via the application of laser
pulses.
Benign tumors of the voice box (e.g., cysts and nodules)
are ideal targets, as they are known to 
respond well to laser treatment~\cite{Shoffel2019}.
Laser application is halted by the operating
physician when visible \textit{blanching} 
(i.e., whitening of the tissue,
which signals thermal
necrosis~\cite{DelSignore2016})
is observed; at this point, the
endoscope is retracted and the 
procedure is concluded.
It is important to note that no tissue is excised
in the course of an office procedure: Necrotized
tumors are left in place, where they spontaneously
\textit{involute}
(i.e., necrotized tissue is re-absorbed and 
eliminated as part of the body's own healing processes)
over the course of a few weeks~\cite{Tibbetts2019}.
\subsection{Challenges of Office Procedures}
While office procedures have been shown to be
generally safe and effective, recent clinical
studies revealed that there exist several
hard-to-reach locations inside the larynx where
laser treatment cannot be 
delivered~\cite{DelSignore2016,Hu2017}.
This issue can be attributed to the limited
dexterity of clinical instruments:
the laser fibers used in office procedures
can only emit light forward in a line-of-sight
fashion, making it hard to reach tumors
that lie off the longitudinal axis of the fiber
(refer to Fig.~\ref{fig:fig-1}).
Physicians can control the laser aiming by 
bending the distal tip of the endoscope; 
however, this does not provide
sufficient anatomical
coverage~\cite{DelSignore2016,Hu2017}.
In recent work, our group characterized the
reachable workspace of laryngeal endoscopes and 
identified the regions within the larynx
where access is problematic~\cite{Chan2021}.
These regions include, among others, the 
two cavities immediately above the vocal folds
and the inferior surface of the vocal folds (see Fig.~\ref{fig:fig-1}).
Patients who develop tumors in one of these
areas are not suitable candidates for office
surgery and have to resort to the traditional
(and more expensive) surgical treatment in the
operating room.
Reachable workspace maps from our earlier study~\cite{Chan2021} are reproduced in this paper
(sec.~\ref{sec:simulation-experiments})
and compared to the reachable workspace
of the new steerable
fiber we propose in the following.
\begin{figure*}
\begin{center}
\includegraphics[width=1\linewidth]{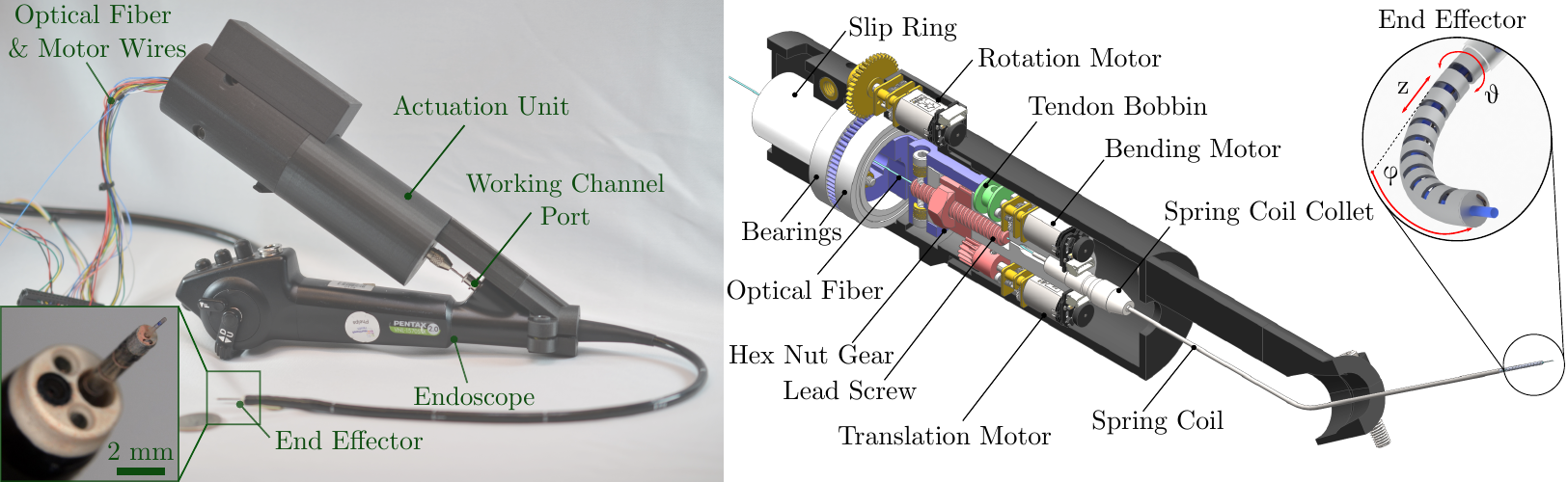}
\caption{Proposed Robotic Steerable Fiber. (Left)
Overview of the device coupled to a hand-held flexible laryngeal endoscope, the Pentax VNL-1570STK
(Pentax Corp., Tokyo, Japan).
The end effector is deployed through the working
channel of the endoscope, and it is controlled by an
actuation unit that mounts on the endoscope handle. 
This design solution enables the operating physician to
hold both the endoscope and the steerable fiber
with a single hand. The weight of the actuation unit
is 250 grams.
(Right) Rendering of the actuation units with callouts
indicating the main components. The inset shows
the end effector of the robot and its three 
degrees of freedom, namely bending, rotation, and
translation.
}
\label{fig:system-overview}
\end{center}
\end{figure*}
\subsection{Design Specifications}
To overcome the kinematic limitations of 
current clinical instruments, and thus amplify a 
physician's reach into the larynx, in this
paper we propose a robotic steering mechanism to
enable the controlled bending of the optical
fibers used for laser delivery.
To guide the design of our device, we used the
following specifications:
\subsubsection{Range of Motion}
Based on our prior analysis of the laryngeal 
    workspace in~\cite{Chan2021}, we anticipate
    that a mechanism able to bend the optical
    fiber by at least
    90\textdegree~would double the
    extent of reachable anatomy and enable access to
    several challenging areas, including those
    immediately above and below the vocal
    folds mentioned earlier.

\subsubsection{Miniaturization} The 
    steering mechanism should be sufficiently
    small to permit deployment of the
    fiber through the working channel of the
    operating endoscope.
    Meeting this specification is important 
    to minimize patient discomfort: a larger 
    mechanism would have to be deployed separately
    from the endoscope, either through the 
    mouth or the other nostril.
    While several miniaturized laser 
    steering mechanisms have been proposed in
    recent years~\cite{York2021,Ferhanoglu2014},
    none of them is sufficiently small to pass
    through the 2 mm lumen of a laryngeal scope.
\subsubsection{Ease of Integration in the Surgical Workflow}
    Introducing the robotic steering mechanism should not
    fundamentally alter the surgical setup or
    workflow. Ideally, physicians should be able 
    to continue to perform the procedure using 
    the instrumentation they are already familiar
    with and a setup as close
    as possible to the one
    illustrated in Fig.~\ref{fig:fig-2}.
\section{System Overview}
\label{sec:Overview}
Our proposed robotic steerable fiber is shown
in Fig.~\ref{fig:system-overview}.
In this section, we provide
an overview of the two main components
of the device, namely the end effector and 
its actuation unit.
\subsection{End Effector}
\label{sec:steerable-fiber}
\begin{figure}
\begin{center}
\includegraphics[width=1\linewidth]{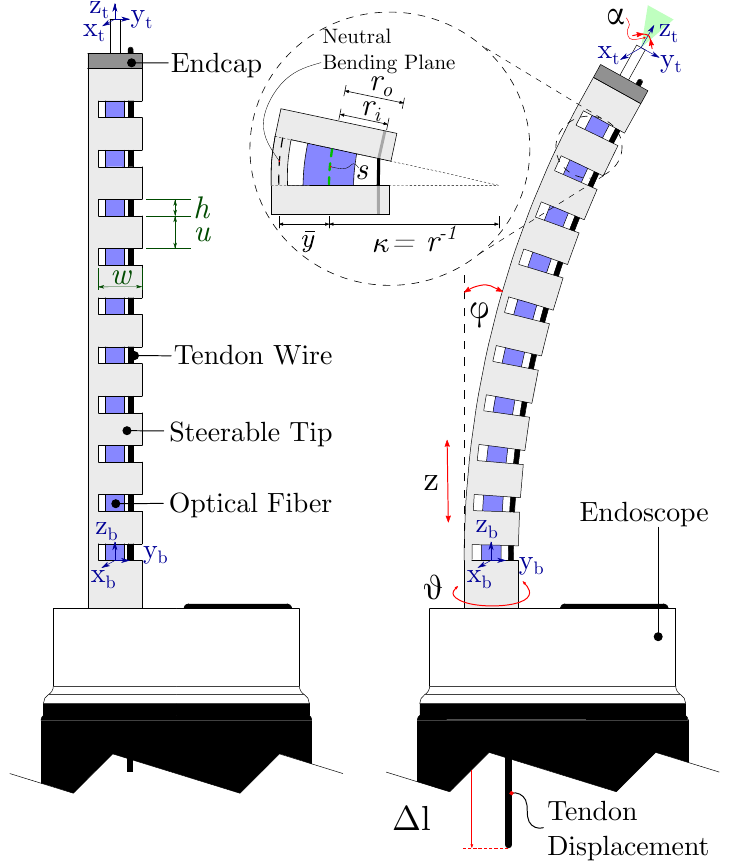}
\caption{The end effector of the robot consists of
an optical fiber installed inside a tendon-actuated
Nickel-Titanium (Nitinol) notched sheath.
The actuation tendon is attached to an aluminum
end cap installed at the distal tip of the sheath.
Applying tension on the tendon makes the notches 
close, creating bending.
When tension is removed, the sheath recovers its
original undistorted configuration.
(Left) Home configuration (no tension on the tendon);
the design parameters of the notched sheath are the notch height $h$, the notch width $w$, and the
spacing $u$ between each pair of notches.
(Right) The end effector provides 
three DoFs, namely tendon displacement
$\Delta l$, axial translation $z$, and
axial rotation $\vartheta$.
Each notch is assumed to bend in the 
shape of constant curvature 
arc, as per~\cite{Swaney2017}.
The laser beam divergence is
$\alpha$ = 40\textdegree. The inset shows the arc parameters that describe the deflection of a single notch, the arc curvature $\kappa$  and the arc length $s$. Variables $r_i$ and
$r_o$ indicate the inner and outer sheath
radius, respectively.
}
\label{fig:fig-4}
\end{center}
\end{figure}

A schematic of the end effector,
illustrating its components and
dimensions, \textcolor{black}{is} 
shown in Fig.~\ref{fig:fig-4}.
It consists of an off-the-shelf optical
fiber installed into a tendon-actuated
Nickel-Titanium (Nitinol) notched continuum
sheath. 
The optical fiber is the FP200ERT
(Thorlabs, Newton, NJ, USA), a multimode fiber
with a diameter of 0.225 mm, capable of withstanding
bending radii as tight as 6 mm~\cite{Zhu2021}.
The distal end of the Nitinol sheath is outfitted with 
an aluminum end cap, which is installed by press-fitting.
The end cap has three
holes (see Fig.~\ref{fig:system-overview}):
the center one, 0.3 mm in diameter, holds the optical
fiber, while the other two, 0.2 mm in diameter, are
used as attachment points for the actuation tendon.
Fabrication of the end cap was carried out with a 
Super Mini Mill CNC machine (Haas Automation, Oxnard, CA, USA).
The steerable sheath uses the asymmetric notch 
pattern proposed by Swaney and colleagues
in~\cite{Swaney2017}.
Compared to other continuum mechanisms,
this solution facilitates miniaturization, as bending
only necessitates a single pull wire.
A potential disadvantage of asymmetric notches is that
they only permit unidirectional bending, though this is not a
major concern as long as the sheath can also 
be rotated.
The notched sheath is fabricated starting from a solid
Nitinol tube wherein notches are then cut with a femtosecond laser.
To build our prototype, we purchased Nitinol tubing
from Johnson Matthey (West Chester, PA, USA), while
laser cutting was outsourced to Pulse Systems 
(Concord, CA, USA).
Custom-sized Nitinol tubing can be
expensive for initial prototyping,
therefore we used the smallest tube diameter available in
stock at the time of purchase that also had
a sufficiently large lumen to accommodate
the optical fiber.
For our prototype shown in Fig.~\ref{fig:system-overview},
we used a Nitinol tube with an outer diameter of 1.1 mm
and a wall thickness of 0.1 mm.
The notch dimensions are illustrated in Fig.~\ref{fig:fig-4}
and their values are listed in Table~\ref{tab:design-params}; 
these are the notch height $h$,
width $w$, and the spacing $u$ between
each pair of consecutive notches.
\begin{table}
\caption{Design Parameters of the Notched Sheath}
\label{tab:design-params}
\begin{center}
\begin{tabular}{ccc}
%\hline
\textbf{Name} & \textbf{Symbol} & \textbf{Value (mm)} \\ \hline
Notch Width   & $w$             & 0.94                \\
Notch Spacing & $u$             & 1.31                \\
Notch Height  & $h$             & 0.19                \\
\hline
\end{tabular}
\end{center}
\end{table}
As we shall see in the following paragraph, the
notch dimensions play directly into the
kinematics of the sheath;
the notch dimensions listed in Table~\ref{tab:design-params}
guarantee the desired minimum 
fiber bending of 90\textdegree, and at the same
time ensure that the bending radius of the device
never goes below 6 mm, which could lead to fiber 
breakage.
To describe the kinematics of the end effector, we use the model from~\cite{Swaney2017}.
This model tracks the sheath's bending
as a function of the tendon displacement
$\Delta l$ (refer to Fig.~\ref{fig:fig-4}).
Briefly, the steerable sheath can be 
considered as an open kinematic chain
composed of a sequence of interleaving 
cut and uncut segments.
Uncut sections are assumed not to undergo 
any deformation; therefore, their contribution
to the kinematics is a simple translation
along the local $z$ axis:
\begin{equation}
%\uvec{e}_3
\mathbf{T}_{\text{uncut}} = e^{	\hat{\zeta}u},~\text{with}~\zeta = [0~0~1~0~0~0]^{\text{T}},
\label{eq:ht-transl}
\end{equation}
where $u$ is the notch spacing.
The 
operator \textasciicircum~in Eq.(\ref{eq:ht-transl})
maps twists from $\mathds{R}^6$ to elements of 
$\mathfrak{se}$(3), i.e. the Lie Algebra of the special Euclidean group $SE(3)$.
Evaluating the exponential on the right-hand side
of the equation yields the corresponding
homogeneous  transformation matrix $\mathbf{T}_{\text{uncut}} \in SE(3)$.
For the notched sections, the kinematics model
in~\cite{Swaney2017} assumes bending
in the shape of a constant curvature arc.
The corresponding homogeneous transformation
matrix can be expressed in terms of the arc 
curvature $\kappa$ and length $s$~ (shown in the inset in Fig.~\ref{fig:fig-4}).
\begin{equation}
    \mathbf{T}_{\text{notch}} = e^{\hat{\mathbf{\xi}} s}~\text{with}~\xi = [0~0~1~0~\kappa~0]^{\text{T}}.
\end{equation}
The notch arc parameters $\kappa$ and $s$ are
related to the tendon displacement $\Delta l$,
as it was shown in~\cite{Swaney2017}:
\begin{equation}
    \kappa \approx \frac{\Delta l}{h (r_i+\bar{y}) - \Delta l \bar{y}},
    ~s = \frac{h}{1 + \bar{y} \kappa}
        \label{eq:kappa-s}
\end{equation}
where $r_i$ is the inner radius of the sheath and 
$\bar{y}$ is the location of the neutral bending
plane with respect to the center axis of the sheath.
This latter quantity can be calculated using the 
relations in~\cite{Swaney2017}.
Finally, the transformation matrix between the base of
the steering section and the tip of the robot 
(i.e., frames {b} and {t} in Fig.~\ref{fig:fig-4}) is 
given by:
\begin{equation}
\mathbf{T}_{\text{robot}} = 
\left(
\prod_{i=1}^{n} \mathbf{T}_{\text{notch}}~\mathbf{T}_{\text{uncut}}\right) \mathbf{T_\text{distal}}
\label{eq:steering-fw-kin}
\end{equation}
where $n$ = 10 is the total number of notches used in our design
and  $\mathbf{T_\text{distal}} \in SE(3)$ represents
an offset along the local $Z$ axis 
that accounts for the presence
of the end cap and the optical fiber at the tip of the
end effector.
The maximum sheath bending angle is achieved when 
all the notches are closed. From simple geometry,
this quantity is given by~\cite{Swaney2017}:
\begin{equation}
    \varphi_{\text{max}} = n\frac{h}{r_o + \bar{y}}.
    \label{eq:theta_max}
\end{equation}
By replacing the parameter values from
Table~\ref{tab:design-params} in the equation above,
and calculating the neutral 
bending plane location $\bar{y}$ using the relations in~\cite{Swaney2017}, it can be verified that
our steerable sheath can bend up to 107.15\textdegree,
thus satisfying the first of the design specifications
listed earlier in sec.~\ref{sec:Related-Work}.
Furthermore, one can use the relations in~\cite{Swaney2017}
to verify that the minimum bending radius of the
sheath is 6.9 mm. 
This ensures that the internal optical fiber 
(which, as noted earlier, can bend down to a radius
of 6 mm) will not break due to excessive bending.
\subsection{Actuation Unit} 
\begin{figure*}
\begin{center}
\includegraphics[width=1\linewidth]{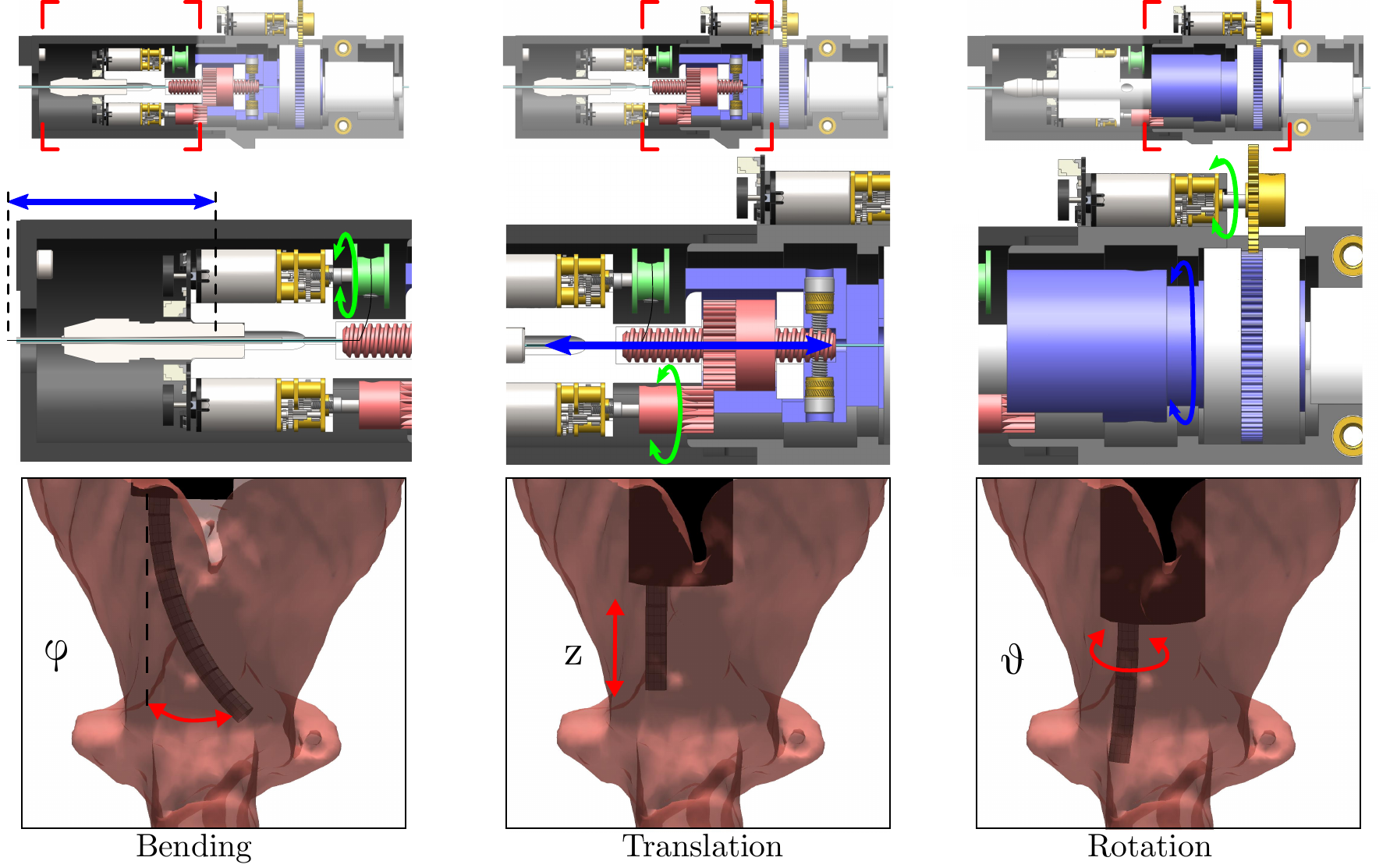}
\caption{
Functional illustration of the actuation unit
(from top: section view of the unit, detailed view of the actuated DoF, and the resulting probe motion within a larynx rendering): (left) rotation of the bending motor actuates the probe in the $\varphi$ direction,
\textcolor{black}{the tendon is routed coaxially through the spring collet and deviates to the tendon bobbin through an adjacent channel (not visible in the cut section)}, 
(middle) actuation of the translation motor rotates the hex nut gear which moves the gear linearly along the fixed lead screw, resulting in a
$z$ movement of the probe, and (right) the distal rotation motor actuates the probe in $\vartheta$.}
\label{fig:functionality}
\end{center}
\end{figure*}
The actuation unit includes three
\textcolor{black}{6 volt}
brushed DC micro-motors
(Pololu Corp., Las Vegas, NV, USA). The
motors and their respective transmission units
are arranged within a 3D-printed
cylindrical housing, 45 mm in diameter \textcolor{black}{and 165 mm in length}.
The unit is rigidly attached to
the endoscope handle via a custom mount, so
that, during a procedure, a physician could hold
both the endoscope and the robot with the same hand.
The weight of the actuation unit is 250 grams.
The operation of the actuation unit is illustrated
in Fig.~\ref{fig:functionality}.
Connection to the end effector uses
a stainless steel
spring coil, the ACT ONE Standard 
(Asahi Intecc, Seto, Japan), which acts as 
a flexible transmission shaft for rotation and
translation.
The hollow lumen of the spring coil is used to pass
the optical fiber and the actuation tendon to the
notched sheath at the end effector.
The spring coil has an outer
diameter of 1.37 mm and an inner diameter of 0.81 mm.
\subsubsection{Bending}
End effector bending is created by pulling
its actuation tendon.
The tendon is wound up on a bobbin (shown in 
green in Fig.~\ref{fig:functionality})
which is directly attached to the shaft of one
of the motors.
\subsubsection{Translation}
Translation is created via a lead screw and nut
mechanism that converts the rotary motion
of another DC motor to linear translation.
\textcolor{black}{The lead screw is hollow to allow the insertion of the laser fiber.}
The encased hex nut moves linearly along the
stationary lead screw, at a rate of 1.60 mm per
turn.
To maintain independence between bending
and linear translation of the end effector,
the bending transmission
travels with the translation mechanism
as to not affect the
tension of the actuation tendon.
The guiding rails built into the design act as
a mechanical limit for the translation
distance.
The total travel distance is 17.0 mm.

\subsubsection{Rotation}
Rotation uses a motor connected to a brass gear,
which interfaces with a custom gear (shown in blue in
Fig.~\ref{fig:functionality}).
The gear ratio between the motor shaft and the
spring coil collet is 7:4.
To prevent twisting in the optical
fiber and entangling of electrical cables, the 
actuation unit uses a through-hole slip ring to
transmit power and signals from the static to the
moving portion of the unit.
\section{Reachable Workspace Characterization}  
\label{sec:simulation-experiments}
\begin{figure}[t]
\begin{center}
\includegraphics[width=1\linewidth]{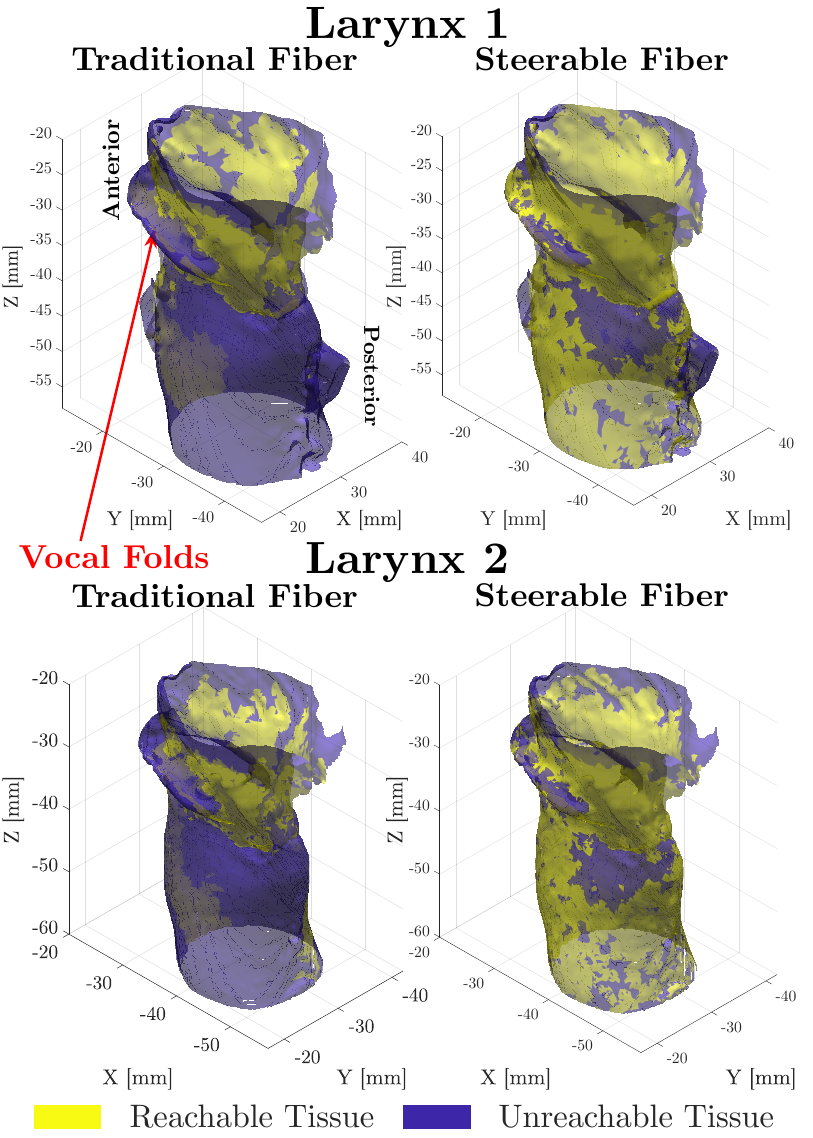}
\caption{Reachable tissue in two different larynx models.
Here, "traditional fiber" refers to the non-steerable
fibers currently in clinical use. "Steerable fiber" 
indicates our new proposed device.
}
\label{simulations}
\end{center}
\end{figure}

This section reports the studies we
performed to characterize the reachable workspace of our
new steerable fiber.
These studies used the same computational simulation
framework we previously developed in~\cite{Chan2021}.
Briefly, we first let a computer program manipulate the
endoscope and the laser fiber in simulation, and
perform an extensive exploration of the reachable
workspace with a sampling-based motion planning
algorithm.
A ray casting procedure is then used to simulate
the application of laser pulses from the fiber tip,
\textcolor{black}{as well as to simulate the field of view of the endoscope camera.}
The output of this process is a three-dimensional
map indicating what tissue %can be accessed
\textcolor{black}{ can simultaneously be reachable by the laser beam and visible by the endoscope camera}
within a given anatomical model.
The endoscope we simulate is the Pentax
VNL-1570STK14 (same make and model shown in
Fig.~\ref{fig:system-overview}).
The kinematics of this endoscope and its range 
of motion were previously determined experimentally 
by our group in~\cite{Chan2021}.
The motion of the steerable wrist is simulated
using the kinematics model previously described
in sec.~\ref{sec:steerable-fiber}.
To simulate the larynx anatomy, we use two
high-resolution three-dimensional larynx models
obtained from micro-tomography scans of
cadavers~\cite{Bailly2018}.
The scans were processed with a MATLAB 
(The Mathworks, Natick, MA, USA) script to 
generate three-dimensional
stereolithography (STL) models.
Reachability maps are calculated via the 
following procedure:
First, we generate 10,000 random endoscope+fiber
configurations using the Rapidly-Exploring Random
Trees (RRT) algorithm.
From each of these reachable configurations,
we run a ray casting algorithm to simulate the
application of laser pulses.
We generate 1,000 virtual rays from the tip of
the fiber in a cone that mimics the laser beam. 
With each of these rays, we use the
M\"{o}ller-Trumbore ray-triangle
Intersection algorithm to detect what faces
of the STL larynx model are visible
in a direct line of sight.
Results are shown in Fig.~\ref{simulations}.
To evaluate the benefits of adding steering to 
laser fibers, we compare our results with 
reachable tissue maps previously calculated by 
our group in~\cite{Chan2021} for clinical
non-steerable laser fibers ("Traditional Fiber" 
in Fig.~\ref{simulations}).
Our new steerable fiber was found to provide
significantly more extensive access to the 
larynx anatomy, compared to traditional clinical
fibers.
Enhanced coverage is observed particularly
in the area surrounding the vocal folds,
which as was noted earlier, can be challenging
to reach with current clinical fibers.
Table~\ref{tab:coveragetissue} shows numerical results,
indicating that being able to bend laser
fibers more than doubles the amount of accessible
tissue. 

%%Result Table
\begin{table}[]
\caption{Tissue Coverage Estimated in Simulation}
\label{tab:coveragetissue}
\setlength{\tabcolsep}{0.7\tabcolsep}% Shrink \tabcolsep by 30%
\centering
\begin{tabular}{ccc}
                  & \textbf{Traditional Fiber}     & \textbf{Steerable Fiber}       \\
                  & \textbf{cm\textsuperscript{2}} & \textbf{cm\textsuperscript{2}} \\ \hline
\textbf{Larynx 1} & 8.99                           & \textcolor{black}{20.65} 
\\
\textbf{Larynx 2} & 8.51                           & \textcolor{black}{19.30}  
\\
\hline

\end{tabular}
\end{table}
\section{Experiments}
\label{sec:experiments}
\subsection{Kinematics Verification}
\begin{figure}
    \centering
    \includegraphics{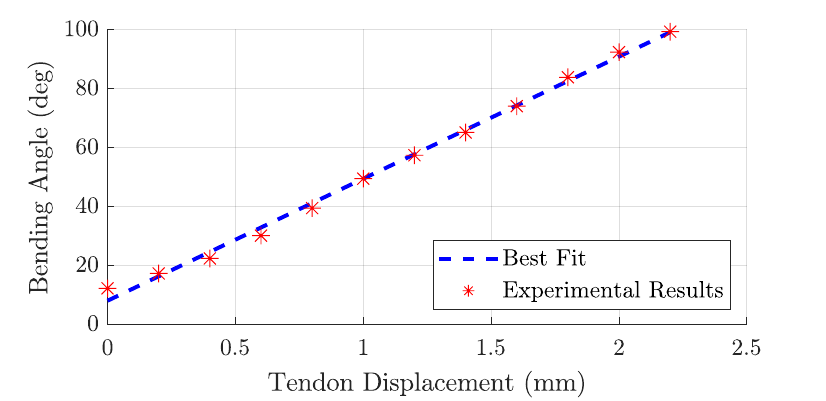}
    \caption{Results of the Kinematics Verification Experiments. 
    \textcolor{black}{The steerable sheath presents an initial pre-curvature of approximately 10\textdegree, which we attribute to 
    heating during the laser cutting process.}
    }
    \label{fig:bending_char}
\end{figure}
To verify that the robot can be controlled using the
kinematic model~\cite{Swaney2017}, we performed 
an experiment wherein the actuation tendon was pulled
in 0.2 mm increments until all the 
notches were observed to be fully closed.
For tendon pulling, we used a manually-actuated
linear slider (Velmex, Bloomfield, NY, USA),
which has a resolution of 0.01 mm.
After each increment, we took a photograph of the 
end effector with a digital single-lens reflex camera 
outfitted with a macro objective (Nikkor 40mmf/2.8 G, Nikon Corporation, Tokyo, Japan) and performed image processing
with a custom MATLAB script to measure the bending angle
$\varphi$.
The pixel resolution of each photograph
was 4928 $\times$ 3624, with a mm-per-pixel ratio of 0.007 mm/pixel. 
The experiment was repeated five times.
\textcolor{black}{After each trial, we reset the position of the steerable fiber without recording the unbending motion.}
The average bending angle for a given amount of tendon displacement $\Delta l$ is displayed in Fig.~\ref{fig:bending_char}. 
The standard deviation among the 5 trials was $<$  1\textdegree for all values of $\Delta l$.
As can be seen in Fig.~\ref{fig:bending_char}, there is a linear relationship between the tendon displacement and the bending angle of the steerable fiber. 
This result is consistent with the modeling work in~\cite{Swaney2017}.
By fitting a linear model, we can map the tendon displacement to the bending angle of the steerable fiber to control its position.
\subsection{Phantom Experiments}
To verify the robot's ability to access hard-to-reach locations
inside the larynx, we performed a set of experiments using a 
3D-printed phantom model of the human larynx.

\begin{figure}
    \centering
    \includegraphics[width=0.8\linewidth]{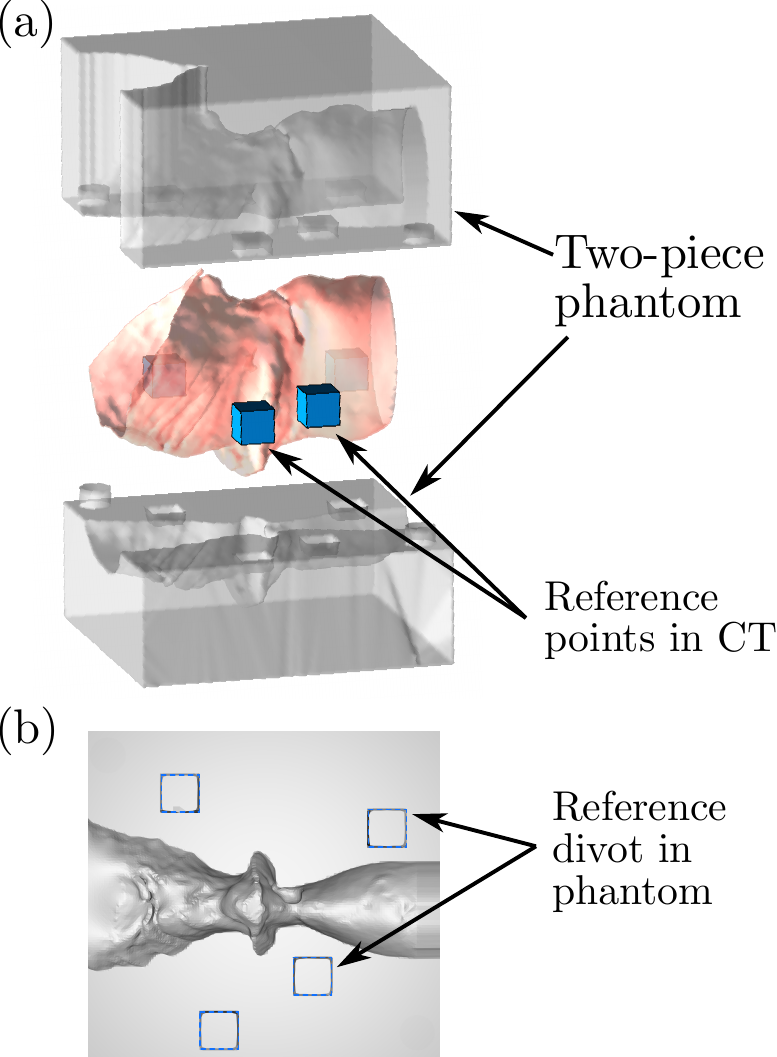}
    \caption{3D printed phantom used in the experiment. (a) An exploded view of the two-piece phantom and larynx mesh model used to generate the phantom. The reference points are shown as blue cubes and were used for registration with the phantom. (b) A planar view of the phantom. The divots (outlined in blue dashes) corresponding to the reference points in the CT scan were localized in the photographs during the experiment.}
    \label{fig:reg}
\end{figure}

\begin{figure}
    \centering
    \includegraphics[width=0.48\textwidth]{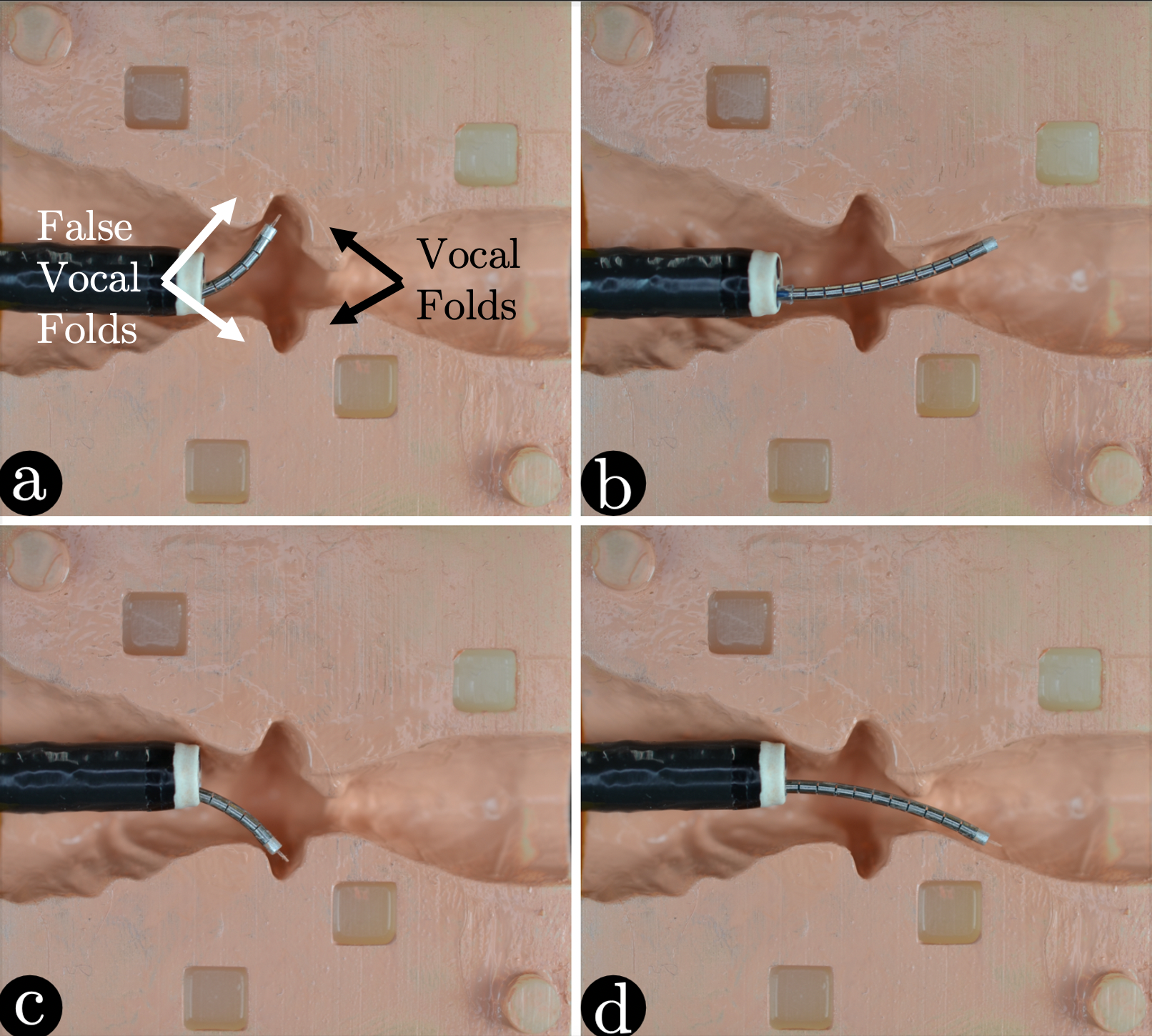}
    \caption{Examples from the phantom experiments can be seen in (a)-(d). The goal was to target positions 
    \textcolor{black}{in between the false and true vocal folds, as well as}
    beneath the right and left vocal folds. Panel (a) and (b) demonstrate the limits for the right vocal fold, and panels (c) and (d) indicate the limits for the left vocal fold.}
    \label{fig:phantom_exp}
\end{figure}

The phantom model, pictured in Fig.~\ref{fig:reg},
was printed with a Form 2 printer (Formlabs, Somerville, MA, USA).
This is the same anatomical model showed earlier in
Fig.~\ref{simulations} and labeled "Larynx 1."
The model was printed as a two-piece phantom, so
that we could open it and take pictures to document
the robot's deployment (see Fig.~\ref{fig:phantom_exp}).

To enable registration between the 3D-printed phantom and 
its STL model, we created four fiducial points,
as shown in Fig.~\ref{fig:reg}.

\begin{figure}
    \centering
    \includegraphics{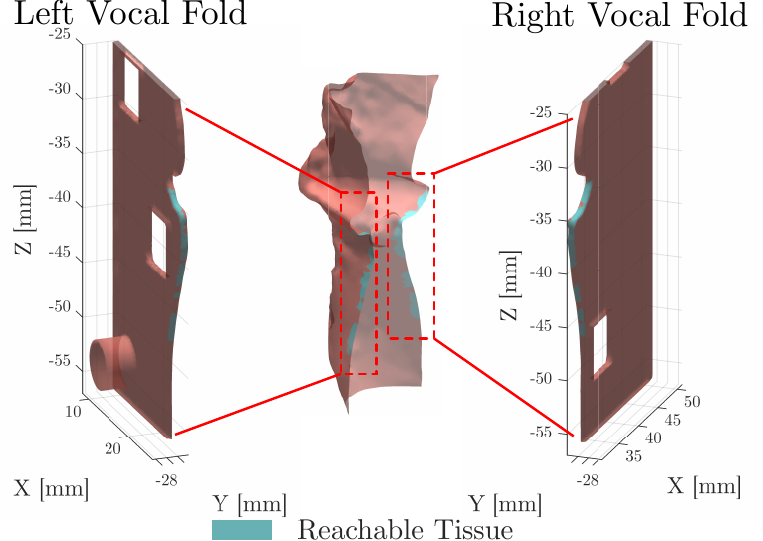}
    \caption{The results of the phantom experiment are shown here for the left and right vocal folds. The steerable probe allows the optical fiber to reach areas inside and beneath the vocal folds.}
    \label{fig:reach_results}
\end{figure}

The experiments were carried out as follows: first, 
the endoscope was deployed into the phantom model so that the steerable fiber probe would navigate in plane with the phantom surface. 
With the endoscope stationary, we deployed the robotic
steerable fiber into the phantom model
\textcolor{black}{and visually aligned it with the surface of the model.}
The robot was controlled using a bench-top
control panel with buttons for each of the
degrees of freedom described earlier in 
sec.~\ref{sec:Overview}.
The steerable fiber was maneuvered to follow the profile
of the tissue surface, scanning from above the vocal folds
to the region immediately below, as shown in
Fig.~\ref{fig:phantom_exp}.
Throughout each experiment, we tracked the location
and orientation of the end effector's tip from images
(using the same camera and lens previously used in the kinematics verification experiments),
and then used the registration between the phantom
and its STL model to project a virtual model of the
robot in the STL space.
This enabled us to use the same ray casting technique 
described earlier in sec. \ref{sec:simulation-experiments} to identify the tissue accessed by the fiber.
Results are shown in Fig.~\ref{fig:reach_results}, and 
they show our robot's ability to reach the regions immediately
above and below the vocal folds, which are 
currently out of reach
with traditional clinical fibers.

\section{Discussion}
\label{sec:Discussion}
Experimental results indicate that the proposed
robotic steerable fiber has the potential to 
substantially enhance surgical access during 
office procedures in the larynx.
There are two particularly hard-to-reach regions
within the larynx, i.e., the set of cavities
immediately above the vocal folds and the 
under-surface of the vocal folds.
Our robot enables access to both of these regions,
as can be seen in Figs.~\ref{simulations}
and~\ref{fig:phantom_exp}.
As of today, patients who present with tumors in
these locations are not eligible for office
surgery and have to resort to traditional surgical
treatment in the operating room.
We expect that the robot we propose in this paper
will help overcome these difficulties and help
move surgeries out of the operating room 
and into the doctor's office.
In separate work~\cite{Zhu2021}, we developed
an optical coupling system to connect our robot
to a surgical laser station, and we demonstrated
the ability of the system to deliver sufficient
laser energy to produce heating and ablation of
tissue.
One limitation of the present work is that 
surgical access was only demonstrated in a limited
number of larynx models. 
Since this is an initial proof-of-principle study, 
inter-patient variability is not a factor that was
accounted for, and it is something that 
we plan to investigate further 
in the future.
Finding high-definition three-dimensional
larynx models has traditionally been difficult
(the larynx anatomy contains tiny
structures that are not straightforward to 
capture with clinical medical imaging protocols);
however, recent years have seen the appearance of a number
of medical image sets, like the one recently
posted by Bailly \textit{et al.} in~\cite{Bailly2018}
that could be used for our purpose.
In future work, we will explore methods to 
optimize the design of the steerable sheath to
further increase anatomical coverage.
More complex designs, such as those 
recently presented in~\cite{Nguyen2022,Pacheco2021},
may allow more comprehensive surgical access
than the simple unidirectional design 
we describe in this paper.
Further studies will be needed to translate
our device to clinical practice.
We designed our robot so that it could be operated
by the operating physician without the need for an
assistant or a supporting arm.
For the purpose of this study, the robot was
controlled with a bench-top control interface. 
While this is useful for initial 
verification and validation, in-office use
dictates placing the user interface on the
actuation unit for single-handed control of the device,
so the operating physician may control the steerable fiber
without disrupting the surgical workflow.
A simple solution would be to integrate control
buttons on the outside shell of the actuation unit,
within reach of the physician's fingers.
Ultimately, we plan to conduct a study where we 
collect inputs from multiple physicians and 
study the ergonomics of different control interfaces.
\section{Conclusion}
\label{sec:Conclusion}
This paper introduced a novel steerable laser fiber for in-office laser surgery in the voice box.
We reported evidence, obtained in simulations and phantom
experiments, documenting the ability of our robot
to substantially enhance surgical access within the 
larynx.
To the best of our knowledge, this paper is the first
report of a surgical robot developed specifically
for office procedures in the larynx.
To design our robot, we explicitly took into account the
surgical setup and workflow of these procedures, and
propose a set of specifications that attempt to 
facilitate the integration of the robot in the surgical setup.
Future work will focus on the development of a user
interface to provide single-handed manipulation.
\bibliographystyle{IEEEtran}
\bibliography{References}

\begin{thebibliography}{10}
\providecommand{\url}[1]{#1}
\csname url@samestyle\endcsname
\providecommand{\newblock}{\relax}
\providecommand{\bibinfo}[2]{#2}
\providecommand{\BIBentrySTDinterwordspacing}{\spaceskip=0pt\relax}
\providecommand{\BIBentryALTinterwordstretchfactor}{4}
\providecommand{\BIBentryALTinterwordspacing}{\spaceskip=\fontdimen2\font plus
\BIBentryALTinterwordstretchfactor\fontdimen3\font minus
  \fontdimen4\font\relax}
\providecommand{\BIBforeignlanguage}[2]{{%
\expandafter\ifx\csname l@#1\endcsname\relax
\typeout{** WARNING: IEEEtran.bst: No hyphenation pattern has been}%
\typeout{** loaded for the language `#1'. Using the pattern for}%
\typeout{** the default language instead.}%
\else
\language=\csname l@#1\endcsname
\fi
#2}}
\providecommand{\BIBdecl}{\relax}
\BIBdecl

\bibitem{Hantzakos2021}
A.~G. Hantzakos and M.~Khan, ``Office laser laryngology: A paradigm shift,''
  \emph{Ear, Nose \& Throat Journal}, vol. 100, no. 1\_suppl, pp. 59S--62S,
  2021.

\bibitem{Chen2021}
S.~Chen, J.~Connors, Y.~Zhang, B.~Wang, D.~Vieira, Y.~Shapira-Galitz,
  D.~Garber, and M.~R. Amin, ``Recurrent respiratory papillomatosis office
  versus operating room: Systematic review and meta-analysis,'' \emph{Annals of
  Otology, Rhinology \& Laryngology}, vol. 130, no.~3, pp. 234--244, 2021.

\bibitem{Chan2021}
I.~A. Chan, J.~F. d'Almeida, A.~J. Chiluisa, T.~L. Carroll, Y.~Liu, and
  L.~Fichera, ``On the merits of using angled fiber tips in office-based laser
  surgery of the vocal folds,'' in \emph{Medical Imaging 2021: Image-Guided
  Procedures, Robotic Interventions, and Modeling}, vol. 11598.\hskip 1em plus
  0.5em minus 0.4em\relax International Society for Optics and Photonics, 2021,
  p. 115981Z.

\bibitem{DelSignore2016}
A.~G. Del~Signore, R.~N. Shah, N.~Gupta, K.~W. Altman, and P.~Woo,
  ``Complications and failures of office-based endoscopic angiolytic laser
  surgery treatment,'' \emph{Journal of Voice}, vol.~30, no.~6, pp. 744--750,
  2016.

\bibitem{Hu2017}
H.-C. Hu, S.-Y. Lin, Y.-T. Hung, and S.-Y. Chang, ``Feasibility and associated
  limitations of office-based laryngeal surgery using carbon dioxide lasers,''
  \emph{JAMA Otolaryngology--Head \& Neck Surgery}, vol. 143, no.~5, pp.
  485--491, 2017.

\bibitem{Bajo2013}
A.~Bajo, L.~M. Dharamsi, J.~L. Netterville, C.~G. Garrett, and N.~Simaan,
  ``Robotic-assisted micro-surgery of the throat: The trans-nasal approach,''
  in \emph{2013 IEEE International Conference on Robotics and
  Automation}.\hskip 1em plus 0.5em minus 0.4em\relax IEEE, 2013, pp. 232--238.

\bibitem{Acemoglu2019}
A.~Acemoglu, N.~Deshpande, J.~Lee, D.~G. Caldwell, and L.~S. Mattos, ``The calm
  system: New generation computer-assisted laser microsurgery,'' in \emph{2019
  19th International Conference on Advanced Robotics (ICAR)}.\hskip 1em plus
  0.5em minus 0.4em\relax IEEE, 2019, pp. 641--646.

\bibitem{Zhao2020}
M.~Zhao, T.~J.~O. Vrielink, A.~A. Kogkas, M.~S. Runciman, D.~S. Elson, and
  G.~P. Mylonas, ``Laryngotors: A novel cable-driven parallel robotic system
  for transoral laser phonosurgery,'' \emph{IEEE Robotics and Automation
  Letters}, vol.~5, no.~2, pp. 1516--1523, 2020.

\bibitem{Mattos2021}
L.~S. Mattos, A.~Acemoglu, A.~Geraldes, A.~Laborai, A.~Schoob, B.~Tamadazte,
  B.~Davies, B.~Wacogne, C.~Pieralli, C.~Barbalata \emph{et~al.}, ``$\mu$ralp
  and beyond: Micro-technologies and systems for robot-assisted endoscopic
  laser microsurgery,'' \emph{Frontiers in Robotics and AI}, p. 240, 2021.

\bibitem{Renevier2017}
R.~Renevier, B.~Tamadazte, K.~Rabenorosoa, L.~Tavernier, and N.~Andreff,
  ``{Endoscopic Laser Surgery: Design, Modeling, and Control},''
  \emph{IEEE/ASME Transactions on Mechatronics}, vol.~22, no.~1, pp. 99--106, 2
  2017.

\bibitem{Kundrat2020}
\BIBentryALTinterwordspacing
D.~Kundrat, R.~Graesslin, A.~Schoob, D.~T. Friedrich, M.~O. Scheithauer, T.~K.
  Hoffmann, T.~Ortmaier, L.~A. Kahrs, and P.~J. Schuler, ``{Preclinical
  Performance Evaluation of a Robotic Endoscope for Non-Contact Laser
  Surgery},'' \emph{Annals of Biomedical Engineering}, pp. 1--16, 8 2020.
  [Online]. Available: \url{https://doi.org/10.1007/s10439-020-02577-y}
\BIBentrySTDinterwordspacing

\bibitem{Fang2021}
G.~Fang, M.~C. Chow, J.~D. Ho, Z.~He, K.~Wang, T.~Ng, J.~K. Tsoi, P.-L. Chan,
  H.-C. Chang, D.~T.-M. Chan \emph{et~al.}, ``Soft robotic manipulator for
  intraoperative mri-guided transoral laser microsurgery,'' \emph{Science
  Robotics}, vol.~6, no.~57, p. eabg5575, 2021.

\bibitem{Zheng2021}
M.~Zheng, N.~Arora, N.~Bhatt, K.~O'Dell, and M.~Johns~III, ``Factors associated
  with tolerance for in-office laryngeal laser procedures,'' \emph{The
  Laryngoscope}, 2021.

\bibitem{Tibbetts2019}
K.~M. Tibbetts and C.~B. Simpson, ``Office-based 532-nanometer pulsed
  potassium-titanyl-phosphate laser procedures in laryngology,''
  \emph{Otolaryngologic Clinics of North America}, vol.~52, no.~3, pp.
  537--557, 2019.

\bibitem{Shoffel2019}
H.~Shoffel-Havakuk, B.~Sadoughi, L.~Sulica, and M.~M. Johns~III, ``In-office
  procedures for the treatment of benign vocal fold lesions in the awake
  patient: a contemporary review,'' \emph{The Laryngoscope}, vol. 129, no.~9,
  pp. 2131--2138, 2019.

\bibitem{York2021}
P.~A. York, R.~Pe{\~n}a, D.~Kent, and R.~J. Wood, ``Microrobotic laser steering
  for minimally invasive surgery,'' \emph{Science Robotics}, vol.~6, no.~50, p.
  eabd5476, 2021.

\bibitem{Ferhanoglu2014}
O.~Ferhanoglu, M.~Yildirim, K.~Subramanian, and A.~Ben-Yakar, ``A 5-mm
  piezo-scanning fiber device for high speed ultrafast laser microsurgery,''
  \emph{Biomedical optics express}, vol.~5, no.~7, pp. 2023--2036, 2014.

\bibitem{Zhu2021}
M.~Zhu, Y.~Shen, A.~J. Chiluisa, J.~Song, L.~Fichera, and Y.~Liu, ``Optical
  fiber coupling system for steerable endoscopic instruments,'' in \emph{2021
  43rd Annual International Conference of the IEEE Engineering in Medicine \&
  Biology Society (EMBC)}.\hskip 1em plus 0.5em minus 0.4em\relax IEEE, 2021,
  pp. 4871--4874.

\bibitem{Swaney2017}
P.~J. Swaney, P.~A. York, H.~B. Gilbert, J.~Burgner-Kahrs, and R.~J. Webster,
  ``{Design, fabrication, and testing of a needle-sized wrist for surgical
  instruments},'' \emph{Journal of Medical Devices, Transactions of the ASME},
  vol.~11, no.~1, p. 014501, 12 2017.

\bibitem{Bailly2018}
L.~Bailly, T.~Cochereau, L.~Orgeas, N.~H. Bernardoni, S.~R. Du~Roscoat,
  A.~Mcleer-Florin, Y.~Robert, X.~Laval, T.~Laurencin, P.~Chaffanjon
  \emph{et~al.}, ``3d multiscale imaging of human vocal folds using synchrotron
  x-ray microtomography in phase retrieval mode,'' \emph{Scientific reports},
  vol.~8, no.~1, pp. 1--20, 2018.

\bibitem{Nguyen2022}
D.~V.~A. Nguyen, C.~Girerd, Q.~Boyer, P.~Rougeot, O.~Lehmann, L.~Tavernier,
  J.~Szewczyk, and K.~Rabenorosoa, ``A hybrid concentric tube robot for
  cholesteatoma laser surgery,'' \emph{IEEE Robotics and Automation Letters},
  vol.~7, no.~1, pp. 462--469, 2022.

\bibitem{Pacheco2021}
N.~E. Pacheco, J.~B. Gafford, M.~A. Atalla, R.~J. Webster~III, and L.~Fichera,
  ``Beyond constant curvature: A new mechanics model for unidirectional
  notched-tube continuum wrists,'' \emph{Journal of Medical Robotics Research},
  vol.~6, no. 01n02, p. 2140004, 2021.

\end{thebibliography}

\end{document}